\documentclass[journal]{IEEEtran}
\usepackage{amsfonts}
\usepackage{amsmath}
\usepackage{graphicx}
\usepackage{xcolor}
\usepackage{amssymb}
\usepackage{booktabs}
\usepackage[sort&compress,numbers]{natbib}
\usepackage[pagebackref=false,colorlinks=true,bookmarks=false]{hyperref}
\begin{document}
%
\title{Segmenting Objects in Day and Night: Edge-Conditioned CNN for Thermal \\Image Semantic Segmentation}
%
%
%

\author{Chenglong Li,
        Wei Xia,
		  Yan Yan,
		  Bin Luo,
        and~Jin Tang
\thanks{C. Li, W. Xia, B. Luo and J. Tang are with School of Computer Science and Technology, Anhui University.}
\thanks{Y. Yan is with Department of Computer Science, Texas State University.}
}

\markboth{IEEE Transactions on xxx}%
{Shell \MakeLowercase{\textit{et al.}}: Bare Demo of IEEEtran.cls for IEEE Journals}

\maketitle

\begin{abstract}
Despite much research progress in image semantic segmentation, it remains challenging under adverse environmental conditions caused by imaging limitations of visible spectrum. 
While thermal infrared cameras have several advantages over cameras for the visible spectrum, such as operating in total darkness, insensitive to illumination variations, robust to shadow effects and strong ability to penetrate haze and smog. 
These advantages of thermal infrared cameras make the segmentation of semantic objects in day and night. 
In this paper we propose a novel network architecture, called edge-conditioned convolutional neural network (EC-CNN), for thermal image semantic segmentation.
Particularly, we elaborately design a gated feature-wise transform layer in EC-CNN to adaptively incorporate edge prior knowledge. 
The whole EC-CNN is end-to-end trained, and can generate high-quality segmentation results with the edge guidance. 
Meanwhile, we also introduce a new benchmark dataset named ``Segment Objects in Day And night''(SODA) for comprehensive evaluations in thermal image semantic segmentation. 
SODA contains over 7,168 manually annotated and synthetically generated thermal images with 20 semantic region labels and from a broad range of viewpoints and scene complexities.
Extensive experiments on SODA demonstrate the effectiveness of the proposed EC-CNN against the state-of-the-art methods.

\end{abstract}


\IEEEpeerreviewmaketitle

\section{Introduction}
The task of image semantic segmentation is to assign a semantic label to every pixel in an image. 
It is a high-level task that paves the way towards scene understanding, whose importance is highlighted by the fact that an increasing number of applications nourish from inferring knowledge from imagery, including autonomous driving, computational photography and augmented reality, to name a few. 
Although remarkable progress has been achieved in the past decades, image semantic segmentation is still considered to be  challenging due to the imaging limitations of visible cameras. 
For example, visible cameras are very sensitive to lighting conditions and become invalid in total darkness. 
Moreover, their imaging quality decrease significantly in adverse environmental conditions, such as rain, haze, and smog.

Thermal infrared cameras can capture infrared radiation (0.75-13um) emitted by subjects with a temperature above absolute zero. 
Thus they can be operated in total darkness, insensitive to illumination variations and robust to shadow effects and also have a strong ability to penetrate haze and smog. 
Moreover, the thermal cameras have recently become affordable and are thus applied to many computer vision tasks, such as visual tracking~\cite{Li16tip,Li18eccv}, object detection~\cite{Hwang15cvpr,Xu17cvpr}, and person Re-ID~\cite{Wu17iccv}. 
Given the potentials of thermal data, it is interesting to investigate the problem of semantic segmentation on thermal images, called thermal image semantic segmentation in this paper. 
However, what is a reasonable baseline algorithm for thermal image semantic segmentation? 
And how to construct a reasonable size and annotated thermal image benchmark for persuasive comparisons of different algorithms?

\begin{figure}[t]
  \centering
  \includegraphics[width=\columnwidth]{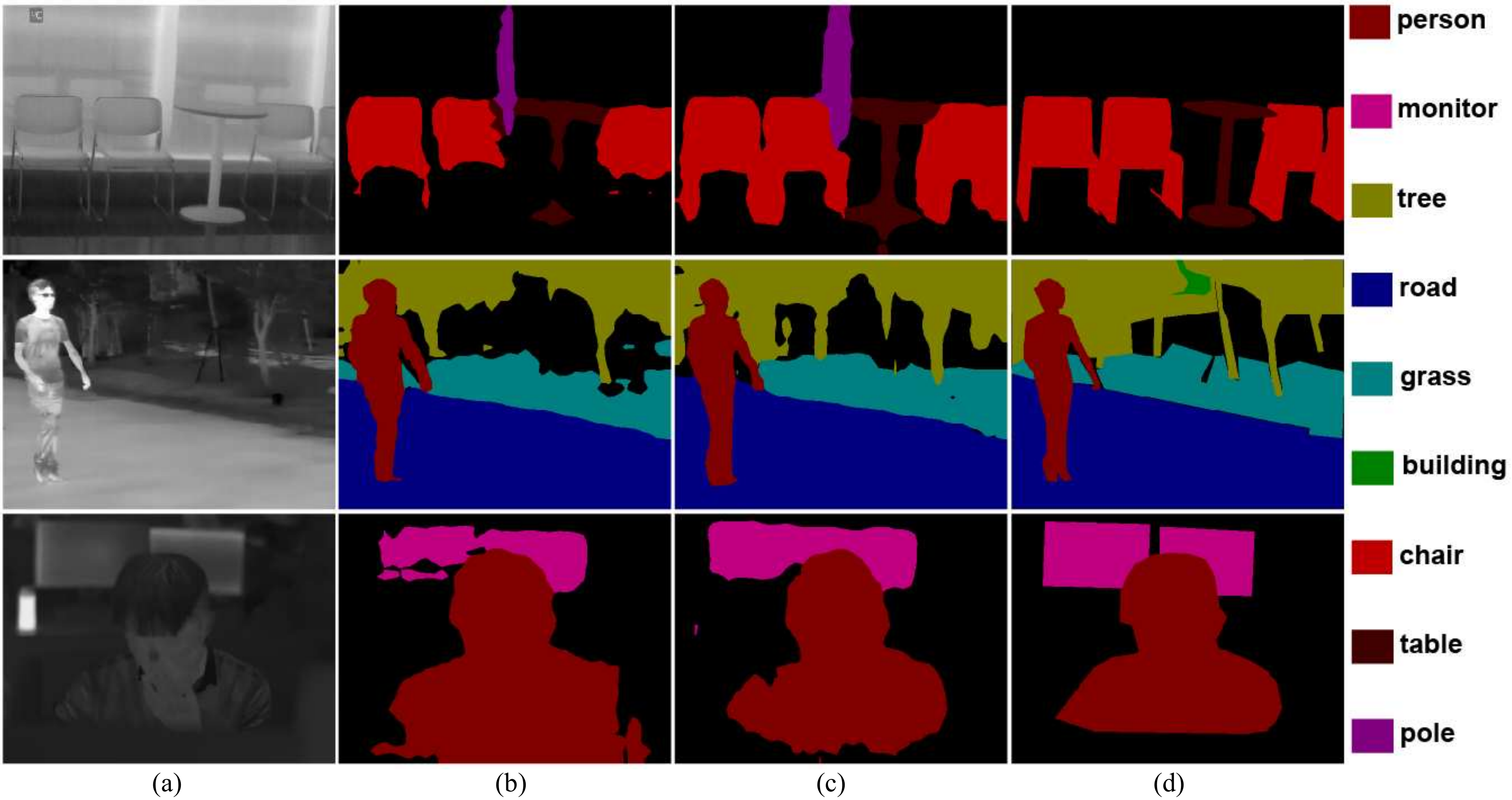} \\
  \caption{Three examples for demonstrating the effectiveness of the proposed EC-CNN. (a) Input thermal images. (b) Results generated by the state-of-the-art network DeepLabv3~\cite{chen2017rethinking}. (c) Results generated by our EC-CNN. (d) Ground truth annotations. }\label{fig::Examples}
\end{figure}

This paper seeks to answer these questions by proposing a powerful baseline algorithm for thermal semantic segmentation, creating a comprehensive and elaborately annotated thermal image dataset and conducting a quantitative benchmark evaluation for different baseline algorithms. 
First, we propose a novel network architecture, called edge-conditioned convolutional neural network (EC-CNN), for thermal image semantic segmentation. 
Despite many advantages on visible spectrum images~\cite{chen2018deeplab,chen2017rethinking}, thermal images are usually with low resolution and contrast, ambiguous object boundaries caused by thermal crossover, and imaging noised introduced by thermal sensors.
Therefore, directly adopting existing techniques for thermal image semantic segmentation is difficult to distinguish different semantic objects accurately, as shown in Fig.~\ref{fig::Examples} (a) and (b). 
To handle these problems, we employ the edge prior knowledge as weakly supervised information to guide thermal image semantic segmentation, and the results are demonstrated in Fig.~\ref{fig::Examples} (c). 
In particular, based on the network DeepLabv3~\cite{chen2017rethinking}, we elaborately design the gated feature-wise transform (GFT) layer in EC-CNN to adaptively embed edge information as the guidance of semantic segmentation.
The major advantages of GFT are as follows: 1) It is parameter-efficient as just a single forward pass through a single network conditioned on the edge prior is needed; 
2) it is able to alleviate effects of noisy edge information via the designed gate scheme;
3) it is able to generate high-quality segmentation results through the adaptive guidance of the edge probability maps; 
and 4) it is generic as it could be applied to other network structures of semantic segmentation. 
Note that we train our EC-CNN in an end-to-end manner.

Second, we establish a standard thermal image benchmark for semantic segmentation. 
There is no public dataset for thermal image semantic segmentation (see Table~\ref{tab:compare_other_datasets} for details), and we contribute such benchmark dataset to facilitate the research. 
Taking advantages of thermal imaging, we are able to Segment any Object in day and night, and thus we call our benchmark dataset SODA for convenience.
SODA contains more than 2,000 manually annotated and 5,000 synthetically generated thermal images with 20 semantic region labels, and covers a broad range of viewpoints and scene complexities.
Particularly, to save the tedious pixel-wise annotation cost, we annotate a small set of thermal images and generate a large set of thermal images using the image-to-image technique~\cite{zhang2019synthetic,wang2018high}.

Finally, we conduct extensive experiments on the SODA dataset. 
To show the advance of our EC-CNN, we include 8 state-of-the-art methods of image semantic segmentation and 3 evaluation metrics (i.e., mean IoU, pixel accuracy and runtime) in our SODA benchmark for persuasive comparisons of different algorithms.
With the deep analysis of these comparison results, we provide some insights and potential future research directions for thermal image segmentation.  

To the best of our knowledge, it is the first work to investigate the problem of thermal image semantic segmentation. Our contributions are summed up as follows.
\begin{itemize}
\item We propose a novel neural network EC-CNN for thermal image semantic segmentation. 
The EC-CNN is able to adaptively embed the edge priors and is thus able to boost the quality of segmentation results.

\item To embed edge prior knowledge effectively, we design a gated feature-wise transform layer to incorporate hierarchical edge maps in an adaptive manner.
The designed layer can perform effective edge embedding while alleviating effects of noises introduced by edge extraction network.

\item We create a new image benchmark dataset SODA for thermal image semantic segmentation. SODA contains over 7,000 thermal images with pixel-wise annotations and 20 semantic labels. 
It provides a first evaluation platform for thermal semantic segmentation, and would be available online for free academic usage~\footnote{Thermal Image Semantic Segmentation Dataset:\\http://chenglongli.cn/people/lcl/dataset-code.html}.

\item Extensive experiments on the SODA dataset are carried out to demonstrate the effectiveness of the proposed EC-CNN. 
Through analysis on experimental results, we provide some insights and potential research directions to the field of thermal image semantic segmentation.

\end{itemize}

\begin{figure*}[t]
  \centering
  \includegraphics[width=\textwidth,height=0.35\textheight]{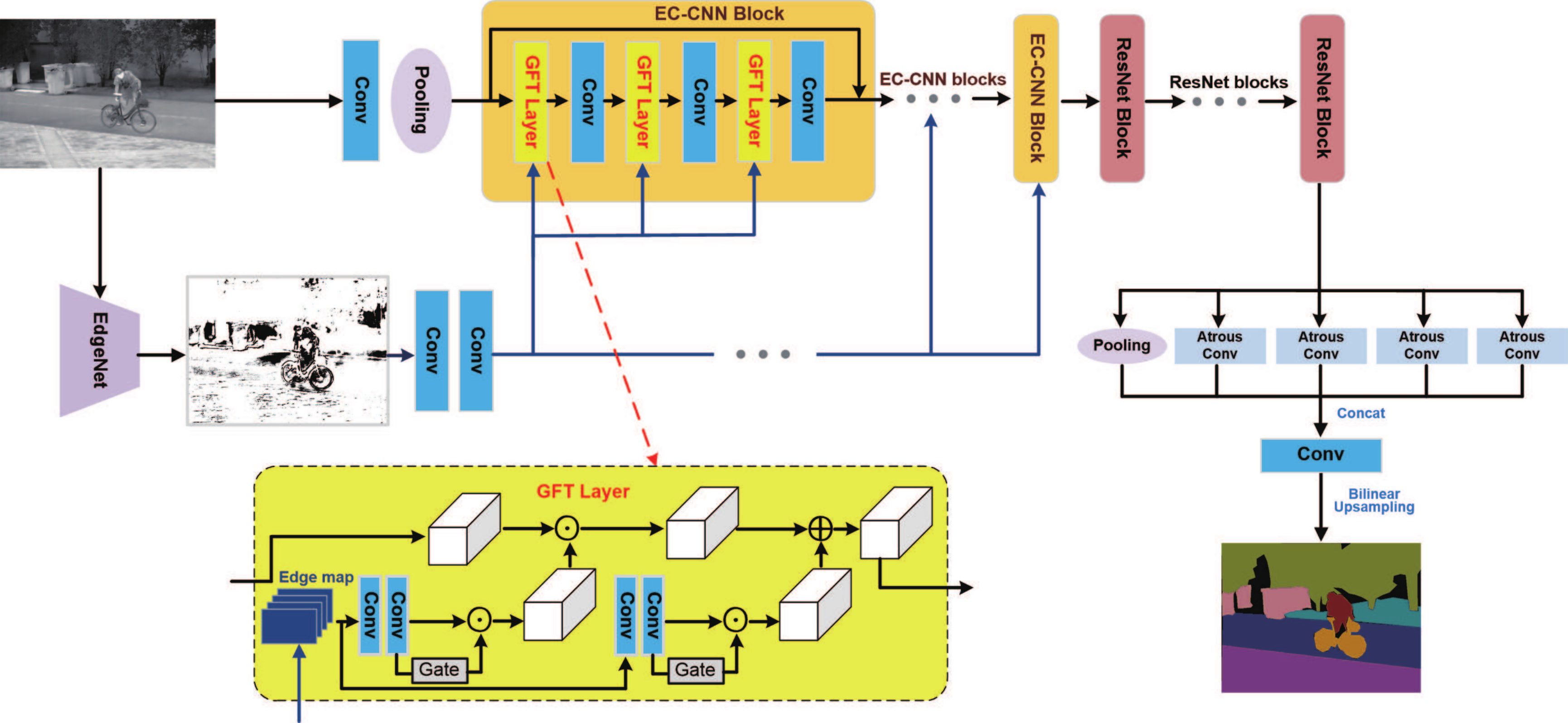} \\
  \caption{Diagram of our EC-CNN architecture. The input image is first processed via an EdgeNet to generate hierarchical edge maps, which are embed into EC-CNN block by inserting the GFT layers to guide the segmentation of input image. Taking the DeepLabv3 as a baseline, we replace some ResNet blocks with EC-CNN blocks for edge prior embedings. } \label{fig::EC-CNN}
\end{figure*}

\section{Related Work} 
This section reviews the most relevant works to ours, including semantic segmentation networks, conditioned networks and adversarial image-to-image translation.

\subsection{Image Semantic Segmentation}
The most successful state-of-the-art deep learning techniques for semantic sementation stem from a common forerunner, the Fully Convolutional Network (FCN)~\cite{long2015fully}. 
FCN replaces the later fully connected layers with convolutional ones to output fully resolution maps instead of classification scores, and is the first work to show how CNNs can be trained end-to-end for this problem. 
Based on FCN, several types of improvements have been made recently.

The encoder-decoder architectures for semantic segmentation~\cite{badrinarayanan2015segnet,ronneberger2015u} first encode longer range information with the spatial dimension gradually reduced and then decode to recover object details and spatial resolution. 
For example, SegNet~\cite{badrinarayanan2015segnet} learns extra convolutional layers to densify the feature response. 
To encode long-range context while preserving object details, four typical methods including image pyramid~\cite{chen2018deeplab,Chen16cvpr}, CRF~\cite{chen2018deeplab}, spatial pyramid pooling~\cite{chen2018deeplab} and atrous convolution~\cite{chen2017rethinking,wang2018understanding} are developed. 
Chen \emph{et al.}~\cite{chen2018deeplab,Chen16cvpr} directly resize the input for several scales and fuse the features from all scales. 
To overcome the poor localization property of CNNs, the DenseCRF~\cite{DenseCRF11nips} is combined with the responses at the final CNN layer~\cite{chen2018deeplab}. 
The atrous spatial pyramid pooling is used in DeepLabv2~\cite{chen2018deeplab} to capture multi-scale information via equipping parallel atrous convolutional layers. 
Chen~\emph{et al.}~\cite{chen2017rethinking} design modules which employ atrous convolution in cascade or in parallel to capture multi-scale context by adopting multiple atrous rates, 
while Wang \emph{et al.}~\cite{wang2018understanding} adopt the atrous convolution with hybrid atrous rates within the last two blocks of ResNet for capturing long-range information.

Besides, there are some works to use edge cues for image semantic segmentation~\cite{chen2016semantic,kokkinos2015pushing,FusionNet}.
Chen~\emph{et al.}~\cite{chen2016semantic} propose to replace the fully-connected CRF with domain transform which is a modern edge-preserving filtering method in which the amount of smoothing is controlled by a reference edge map. 
Kokkinos~\cite{kokkinos2015pushing} designs CRF-based method, which intergrates CNN score-maps with pairwise features derived from advanced boundary detection, to more precisely localize semantic segmentation boundaries.
The outputs of edge network are employed to refine semantic segmentation network by adding an edge aware regularization so as to yield spatially consistent and well boundary located results~\cite{FusionNet}.

In comparison with previous works, we propose an edge-conditioned system that takes advantage of object edges to guide thermal image semantic segmentation. 

\subsection{Conditioned Networks}
Conditional feature normalization uses a learned function of some conditions to achieve feature-wise affine transformation in batch normalization~\cite{ioffe2015batch}, 
and its variants are proven to be effective in several computer vision tasks, such as style transfer~\cite{dumoulin2017learned,huang2017arbitrary}, visual question answering~\cite{de2017modulating}, visual reasoning~\cite{perez2017film} and image super-resolution~\cite{DFT18cvpr}. 
Perez~\emph{et al.}~\cite{perez2017film} propose a feature-wise linear modulation layer (FiLM) to exploit linguistic information for visual reasoning. 
%
For the low-level tasks like image super-resolution and semantic segmentation, the spatial information is crucial and thus should be considered in feature modulation. 
Wang~\emph{et al.}~\cite{DFT18cvpr} propose spatial feature transform to incorporate semantic
prior for image super-resolution.
However, noisy information introduced by the priors might affect the performance.
Therefore, we design a gate scheme in the spatial feature transform to adaptively incorporate edge priors in our framework.

\subsection{Adversarial Image-to-image Translation}
Generative Adversarial Networks (GANs)~\cite{goodfellow2014generative} have achieved promising results in several task such as image generation~\cite{denton2015deep}, image inpainting~\cite{perarnau2016invertible}, and representation learning~\cite{salimans2016improved}. 
The conditional variants of GANs~\cite{mirza2014conditional} enable to condition the image generation on a selected input variable, for example, an input image. 
In this case, the task becomes image-to-image translation, and this is the variant we use here. 
The general method of Isola \emph{et al.}~\cite{isola2017image}, pix2pix, was the first GAN-based image-to-image translation work that was not designed for a specific task. 
The architecture is based on an encoder-decoder with skip connections and it is trained using a combination of losses: a conditional adversarial loss~\cite{goodfellow2014generative} and a loss that maps the generated image close to the corresponding target image. 
This method achieves excellent results, but loss the details of object delineation, in other words, failure for high-resolution image generation task. 
In order to overcome this limitation,~\cite{wang2018high} extended this model to provide a new adversarial learning objective together with new multi-scale generator and discriminator architectures, which are more visually appealing than those computed by previous methods. 
We use it to create synthetic training data for thermal image semantic segmentation, it is much easier to create semantic segmentation datasets for desired thermal image scenarios than to generate training images.

\begin{figure*}[t]
  \centering
  \includegraphics[width=0.9\textwidth]{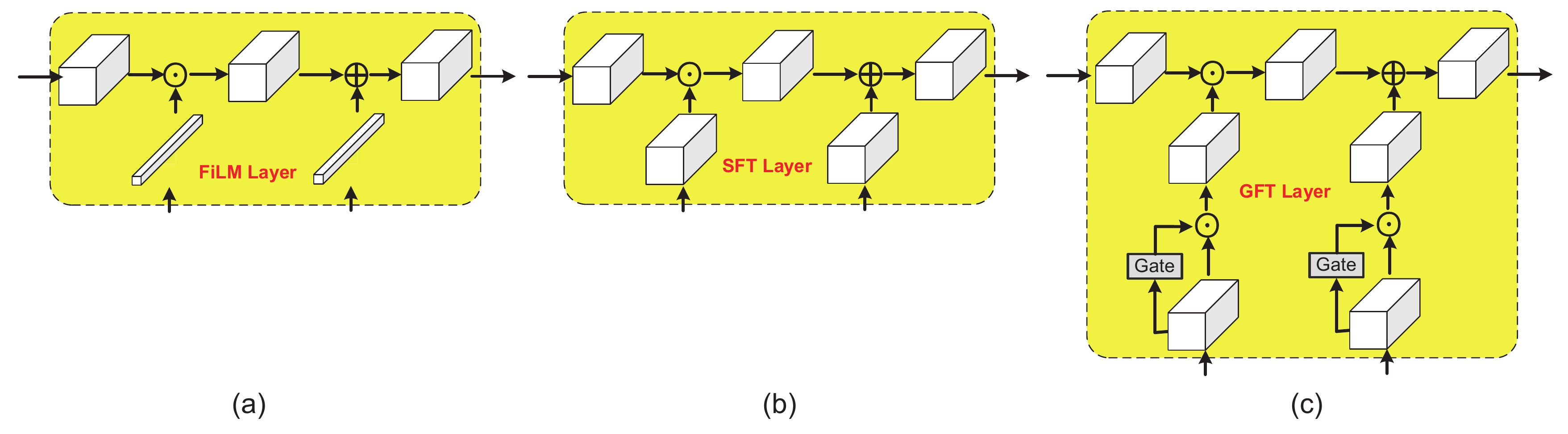} \\
  \caption{Structures of FiLM~\cite{perez2017film}, SFT~\cite{DFT18cvpr} and our GFT shown in (a), (b) and (c) respectively.}\label{fig::film_sft_gft}
\end{figure*}

\section{Edge-Conditioned CNN}
To utilize the edge prior sufficiently from thermal images for semantic segmentation, we design an Edge-Conditioned Convolutional Neural Network (EC-CNN), and this section will present the details of our EC-CNN.

\subsection{EC-CNN Architecture}
As shown in Fig~\ref{fig::EC-CNN}, our EC-CNN consists of two streams, including an edge detection network (EdgeNet) and a novel semantic segmentation network. 

\begin{figure}[t]
  \centering
  \includegraphics[width=\columnwidth,height=0.16\textheight]{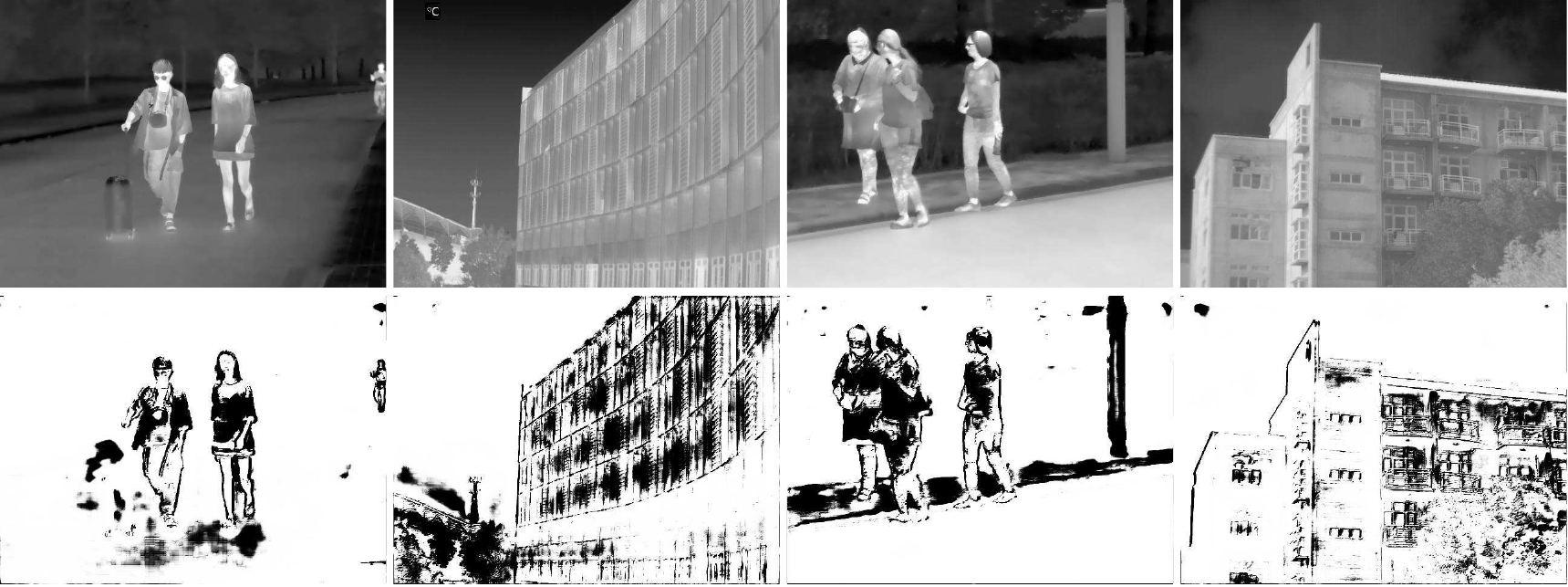} \\
  \caption{The edge results trained on RGB-based dataset}\label{fig::show_edge}
\end{figure}

Despite many advantages on visible spectrum images~\cite{chen2018deeplab,chen2017rethinking}, thermal images are usually with low resolution and contrast, ambiguous object boundaries caused by thermal crossover, and imaging noised introduced by thermal sensors.
To handle this problem, we aim to utilize hierarchical edge features to guide thermal image segmentation and thus equip the EdgeNet in our network.
These edge priors are beneficial for guiding the results of thermal semantic segmentation to possess good object boundaries.
Specifically, we employ the HED (Holistically-nested Edge Detection)~\cite{xie2015holistically} as our EdgeNet to generate high-quality edge features. 
Note that there is no thermal image dataset with ground truth edge annotations.
Therefore, we only train HED using a RGB image dataset, BSD500~\cite{martin2004learning}. 
In spite of using RGB images as training samples, the high-quality edge results of thermal images are also obtained mainly as RGB and thermal images are able to share low-level feature representations, as shown in Fig.~\ref{fig::show_edge}.
%

The proposed semantic segmentation network bases on a state-of-the-art semantic segmentation network, i.e., DeepLabv3~\cite{chen2017rethinking}.
We briefly review the DeepLabv3 here for clarity.
DeepLabv3 takes the structure of the ResNet~\cite{he2016deep} as its feature extractor.
To enable different-level context aggregation and retain more spatial information, it integrates the dilated/atrous convolutions with different rates in the network. 
To handle the problem of segmenting objects at multiple scales, it employs atrous convolution in cascade or in parallel to capture multi-scale context by adopting multiple atrous rates.
A atrous spatial pyramid pooling module is incorporated to probe convolutional features at multiple scales, which encode global context and further boost performance.

To deal with the challenges of low resolution and contrast, and ambiguous object boundaries caused by thermal crossover, we embed edge feature maps generated by  EdgeNet into DeepLabv3.
To this end, we design a EC-CNN block by inserting some {\bf gated feature-wise transform} (GFT) layers, as shown in Fig.~\ref{fig::EC-CNN}, and integrate it into DeepLabv3 so that the edge priors could be deployed to guide thermal image semantic segmentation.  
The GFT layer consists of three components, i.e., {\bf feature modulation}, {\bf information control}, and {\bf feature-wise transform}. 

Since the edge maps generated by EdgeNet have different resolutions and channel numbers from the feature maps of DeepLabv3, we can not utilize the edge priors directly.
To handle this problem, we design a module of feature modulation to transform the resolution and channel number of edge maps into the same with the feature maps.
The feature modulation includes two convolution layers and the hyperparams of them are adaptively set according to the depth of feature maps. 
The edge prior information generated from EdgeNet often includes many noises, and direct use of them might degrade the performance.
Therefore, we design an information control module using a gate scheme to adaptively incorporate edge prior information.
As for the module of feature-wise transform, we use it to embed edge maps into the learning of deep features by applying a spatially affine transformation to each intermediate feature map in DeepLabv3.
Next, we discuss the GFT layer in details.

\begin{figure}[t]
  \centering
  \includegraphics[width=0.45\textwidth]{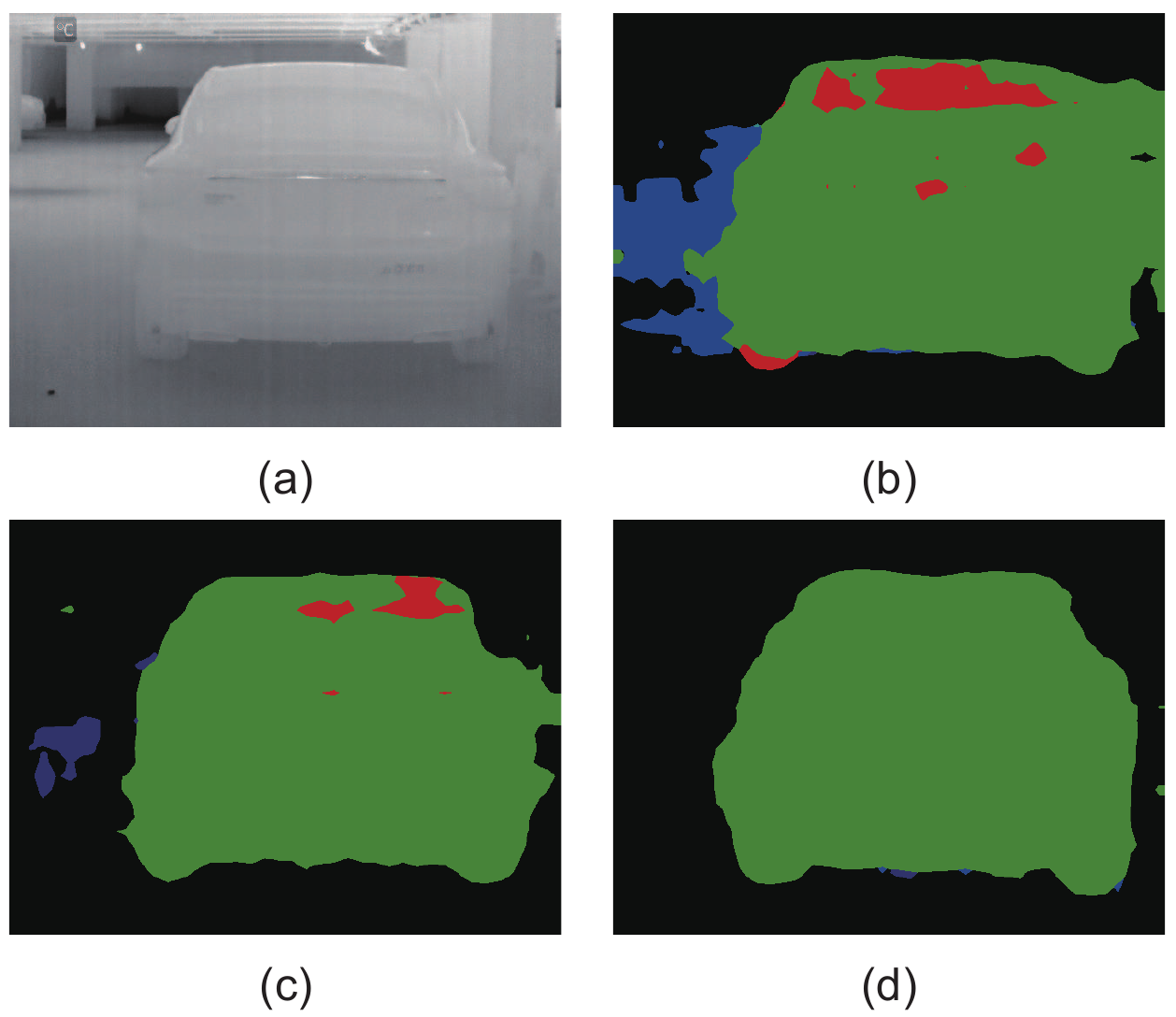} \\
  \caption{Illustration of the proposed components. (a) Input image. (b) Result of DeepLabv3~\cite{chen2017rethinking}. (c) Result of our method without the gate scheme. (d) Result of our method. The results show the effectiveness of the components in our method.}\label{fig::v3_sft_gft_compare}
\end{figure}

\begin{figure}[t]
  \centering
  \includegraphics[width=\columnwidth]{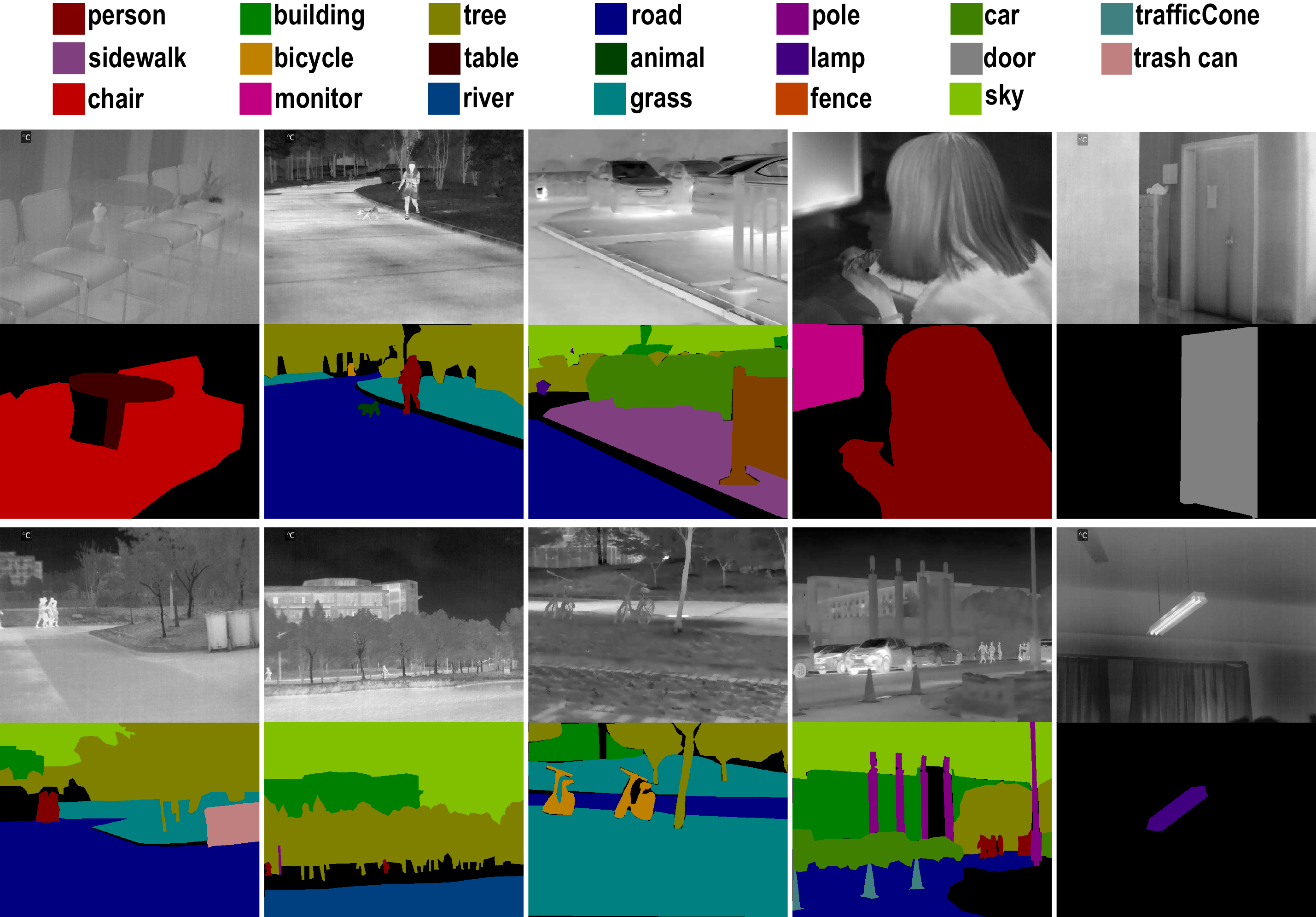} \\
  \caption{Examples of thermal images and their corresponding ground truth annotations from the SODA dataset. }\label{fig::DatasetExamples}
\end{figure}

\begin{figure}[t]
  \centering
  \includegraphics[width=\columnwidth]{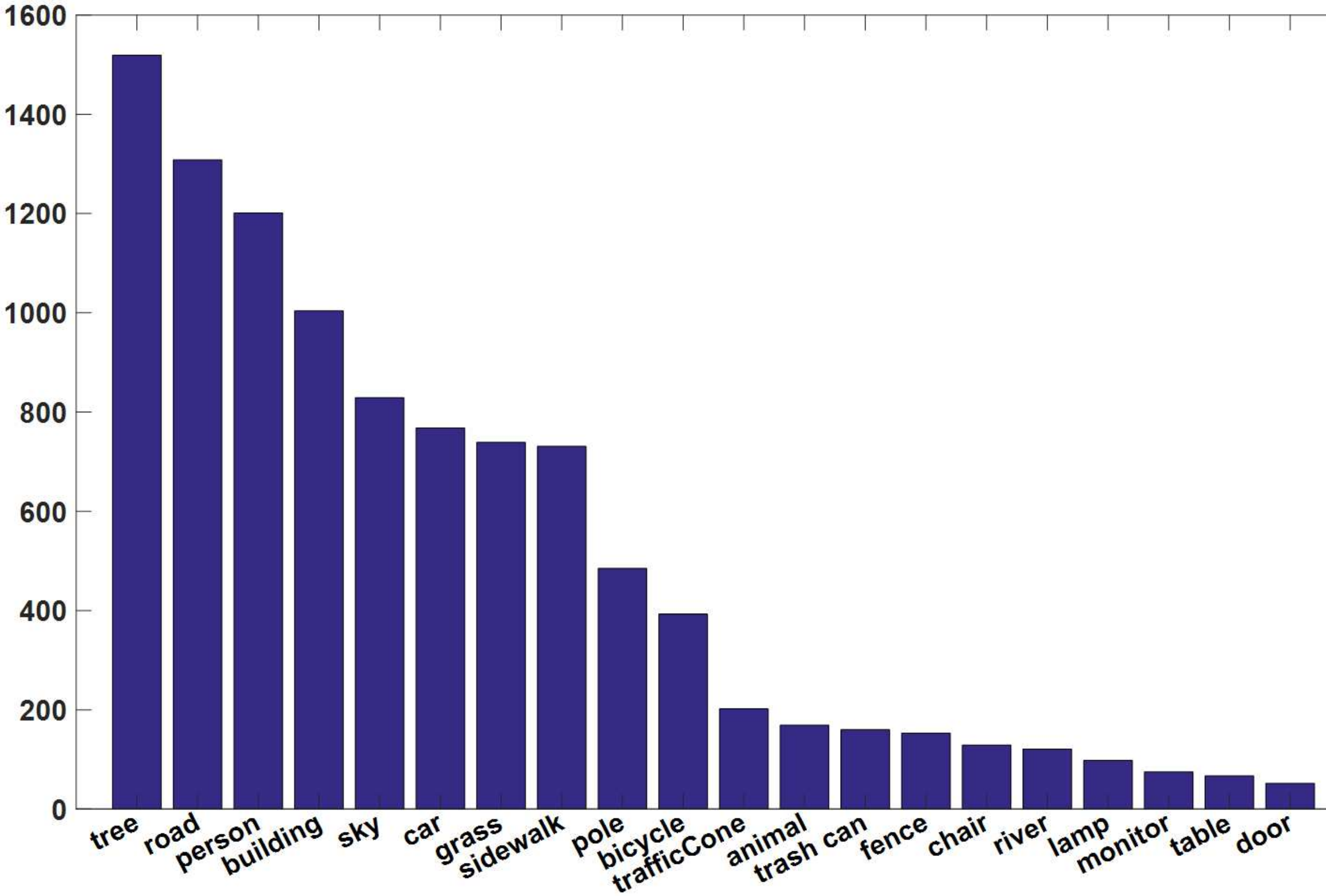} \\
  \caption{Image number distribution of all semantic labels on the SODA dataset. }\label{fig::DatasetStatistics}
\end{figure}

\begin{figure}[t]
  \centering
  \includegraphics[width=0.45\textwidth]{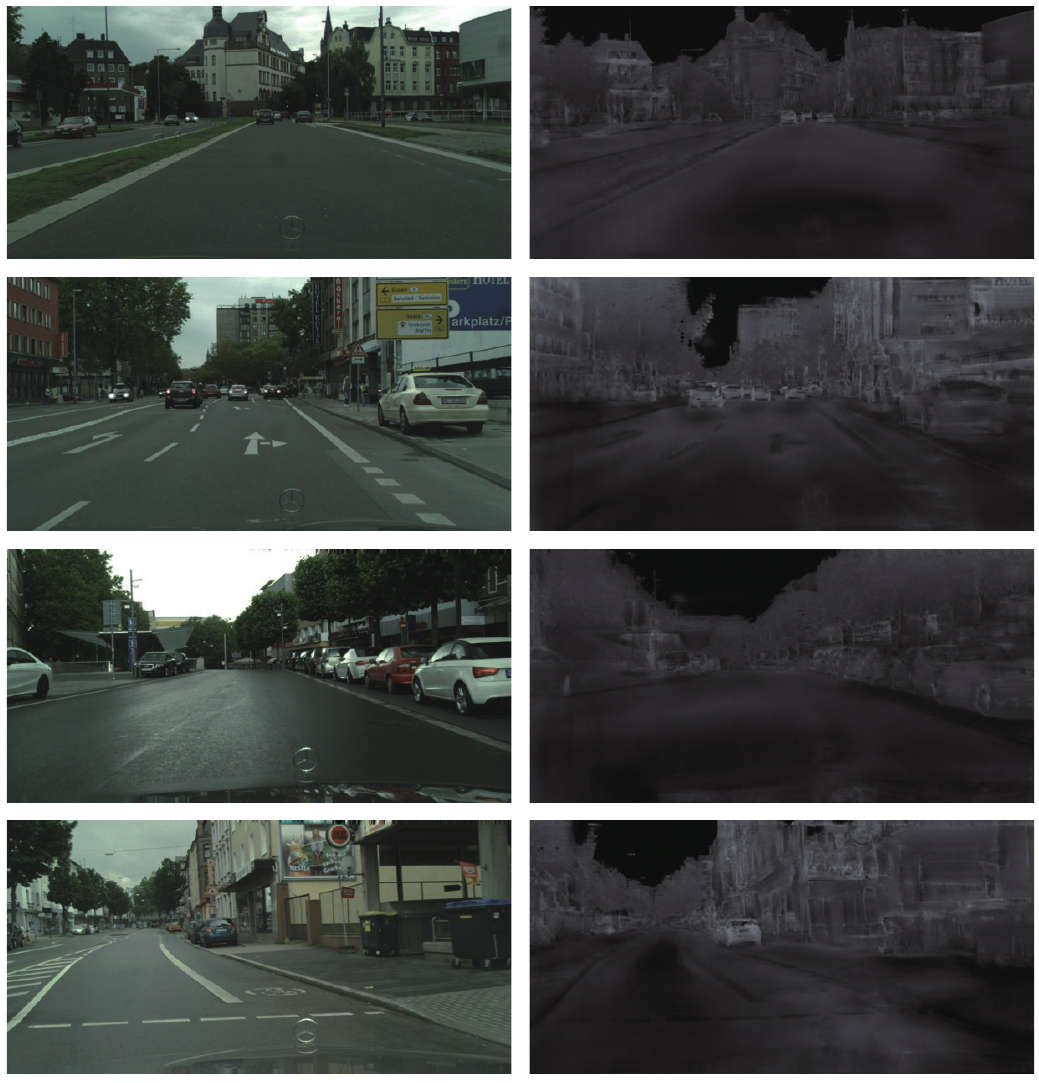} \\
  \caption{The translate TIR of Cityscapes. }\label{fig::cityscapes}
\end{figure}

\subsection{GFT Layer}
\label{body::eccnn block}
Our EC-CNN is inspired by previous studies on feature normalization~\cite{ioffe2015batch,dumoulin2017learned}. 
Batch normalization (BN) is a widely used technique to ease network training by normalizing features statistics~\cite{ioffe2015batch}. 
Conditional Normalization (CN) applies a learned function of some conditions to replace parameters for feature-wise affine transformation in BN. 
Some variants of CN have been proven to be highly effective in image style transfer~\cite{dumoulin2017learned,huang2017arbitrary}, visual question answering~\cite{de2017modulating} and visual reasoning~\cite{Perez17arXiv}.
Perez \emph{et al.}~\cite{perez2017film} develop a feature-wise linear modulation layer (FiLM), to exploit linguistic information for visual reasoning. 
FiLM is a relevant technique to us, and thus we briefly introduce it first.

A FiLM layer applies a feature-wise affine transformation (i.e., the shifting and scaling coefficients) to condition prior knowledge on feature maps for more robust feature learning. 
The structure of FiLM is shown in Fig.~\ref{fig::film_sft_gft} (a). 
We denote that $\mathbf{x}$ is an input of FiLM layer, $\mathbf{z}$ is a conditioning input, and $\gamma$ and $\beta$ are $\mathbf{z}$-dependent functions that represents the scaling and shifting vectors respectively.
The formulation of FiLM can be written as follows:
\begin{equation}
\label{edge:FILM}
FiLM(\mathbf{x}) = \gamma(\mathbf{z})\times \mathbf{x} + \beta(\mathbf{z}),
\end{equation}
where the dimensions of $\gamma(\mathbf{z})$ and $\beta(\mathbf{z})$ are equal to the channel number of $\mathbf{z}$, i.e., each channel of $\mathbf{z}$ is mapped into a scaling parameter and a shifting parameter.

Nonetheless, FiLM cannot handle the conditions with spatial information since it maps each channel of input maps into a single scaling/shifting parameter. 
The spatial information is crucial for low-level tasks like image super-resolution, semantic segmentation, since these tasks usually require adaptive processing at different spatial locations of an image. 
Therefore, Wang~\emph{et al.} propose a spatial feature transform (SFT) to incorporate spatial priors for image super-resolution.
SFT is formulated as follows:
\begin{equation}
\label{edge:sft}
SFT(\mathbf{x}) = \gamma(\mathbf{z})\odot \mathbf{x} \oplus \beta(\mathbf{z})
\end{equation}
where $\odot$ and $\oplus$ denote the element-wise product and sum respectively.
Different from FiLM, the dimensions of $\gamma(\mathbf{z})$ and $\beta(\mathbf{z})$ are same with $\mathbf{z}$, i.e., the input $\mathbf{z}$ is mapped into $\mathbf{z}$-dependent scaling and shifting tensors.
The structure of SFT is shown in Fig~\ref{fig::film_sft_gft} (b). 

We aim to employ SFT to embed the edge priors to guide thermal image semantic segmentation.
However, the EdgeNet often introduce some noisy edges and directly using these edges as guidance would affect the performance of semantic segmentation.
To handle this problem, we propose a new layer called gated feature-wise transform (GFT) to achieve adaptive incorporation of edge priors.
The effectiveness of the proposed GFT is demonstrated in Fig.~\ref{fig::v3_sft_gft_compare}.
As discussed above, GFT includes three modules of feature modulation, information control, and feature-wise transform.
The information control module is used to adaptively incorporate edge prior information, and we propose to use a gate scheme to realize it.
The structure of our GFT is shown in Fig.~\ref{fig::film_sft_gft} (c). 
Specifically, we formulate the information control module as follows:
\begin{equation}
\begin{aligned}
\label{edge:gate} 
&(\gamma^*(\hat{\bf z}), \beta^*(\hat{\bf z})) = \sigma(\gamma(\hat{\bf z}), \beta(\hat{\bf z})),\\
&\hat{\gamma}(\hat{\bf z}) = \gamma(\hat{\bf z}) \odot \gamma^*(\hat{\bf z}),  \\
&\hat{\beta}(\hat{\bf z}) = \beta(\hat{\bf z}) \odot \beta^*(\hat{\bf z}),
\end{aligned}
\end{equation}
where $\sigma$ is the sigmoid function, and $\hat{\bf z}$ is the output of the feature modulation module with the input of ${\bf z}$.
Therefore, the GFT can be formulated as follows:
\begin{equation}
\begin{aligned}
\label{edge:gft}
GFT({\bf x}) = \hat{\gamma}(\hat{\bf z}) \odot {\bf x} \oplus \hat{\beta}(\hat{\bf z}), \\
\end{aligned}
\end{equation}

Fig.~\ref{fig::EC-CNN} shows the detailed structure of the GFT layer. 
%

%



\subsection{Loss Function}
We apply the cross entropy loss for semantic segmentation as follows:
\begin{equation}
\label{edge:loss}
\mathcal{L} = -\frac{1}{n} \sum_i^n[ y_i\ln G_{\theta}({\bf x}_i|{\bf z}) + (1 - y_i) \ln(1- G_{\theta}({\bf x}_i|{\bf z}))]
\end{equation}
where $G_{\theta}({\bf x}_i|{\bf z})$ is the label assignment probability at pixel $i$ with the parameter $\theta$ in our EC-CNN, and $y_i$ is the ground truth label.

\subsection{Network Training}
As for EdgeNet, we first initialize its parameters using the pre-trained VGG-16-Net model on the ImageNet dataset, and then fine-tuned it using the BSD500 dataset~\cite{martin2004learning} to generate hierarchical edge maps, and keep its parameters unchanged in the training procedure of EC-CNN.
Although the model is trained on RGB data, the high-quality edge results of
thermal images are also obtained mainly as RGB and thermal
images are able to share low-level feature representations as discussed above.
To effectively learn the parameters of EC-CNN using our SODA dataset, we adopt a progressive learning algorithm.
First, we load the network parameters pre-trained on the ImageNet dataset as the initial parameters of EC-CNN.
Second, we use the synthetic thermal subset to further train EC-CNN. 
Finally, we fine-tune our network by the real thermal subset.
Note that the labels of Cityscapes dataset are different from our real subset, thus we discard the parameters of the last layer after the second stage.

\begin{figure*}[t]
  \centering
  \includegraphics[width=0.9\textwidth,height=0.45\textheight]{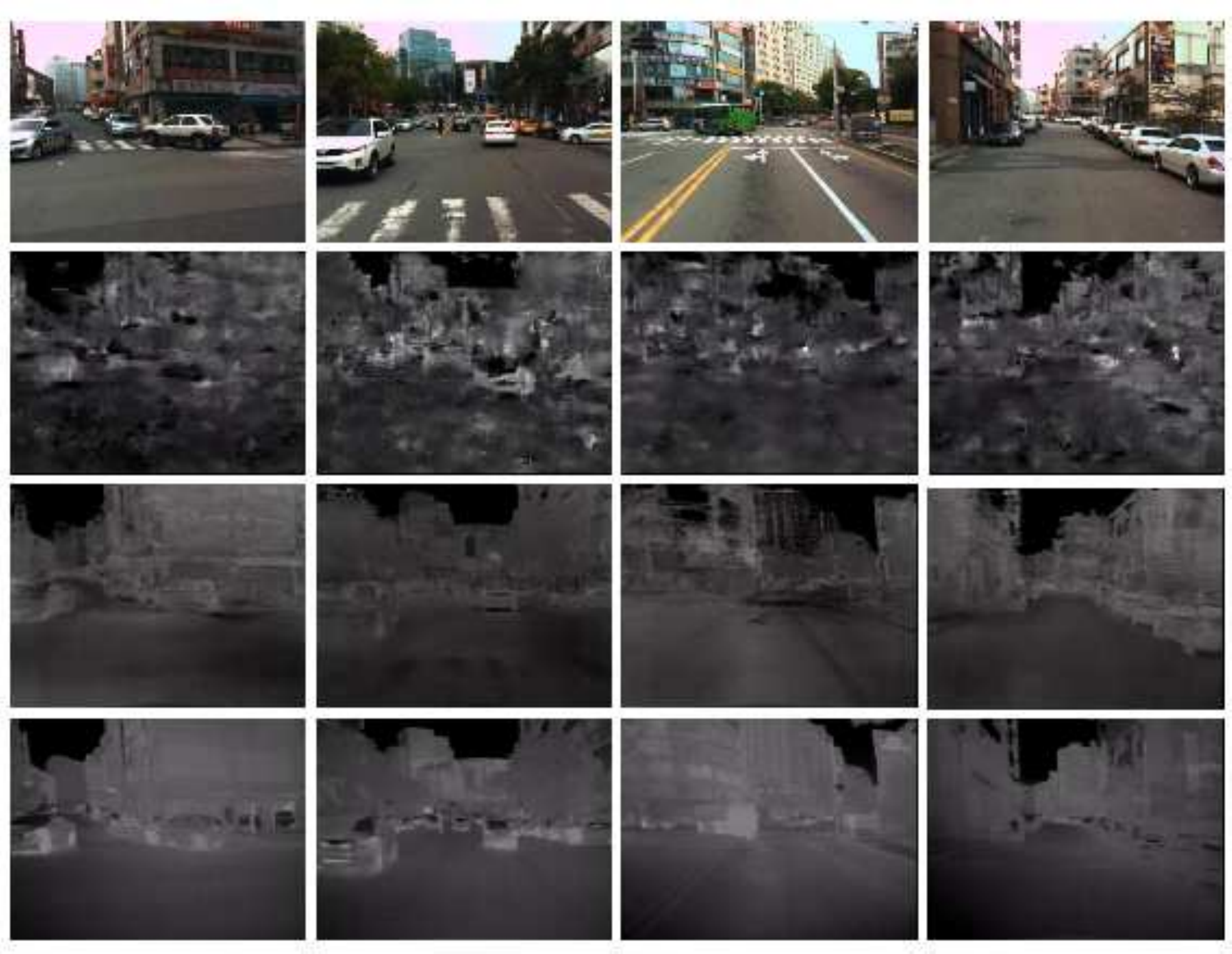} \\
  \caption{Results for the two image translation methods : pix2pix and pix2pixHD. The first row is input image, the second row is pix2pix, the third row is pix2pixHD, the finally row is groundtruth.}\label{fig::pix2pixHD_pix2pix_demo}
\end{figure*}

\section{SODA Dataset}
This section will introduce the benchmark dataset namely ``Segment Objects in Day And night (SODA)''. 
Particularly, our SODA includes 2,168 annotated images and 5,000 synthetically generated thermal images.
The former is captured in real scenes and environments using thermal camera, but the manual annotations are time consuming and also easily include noises due to low contrast and resolution of thermal images.
To handle these problems, we employ some well-annotated RGB images to generate synthetic thermal images.
We first present the details of real part of SODA and then synthetic part.

\subsection{Real Subset}
Our SODA dataset is captured under a FLIR thermal camera (SC620) from a broad range of viewpoints and scene complexities, and contains 2,165 elaborately annotated thermal images with 20 semantic region labels, 
including person, building, tree, road, pole, grass, door, table, chair, car, bicycle, lamp, monitor, trafficCone, trash can, animal, fence, sky, river, sidewalk, and a background label. 
The images collected from the real-world scenarios contain variety appearances with different viewpoints, heavy occlusions, imaging blur and in a wide range of resolutions. 
The open source annotation tool \emph{labelme}, written in Python, is used to annotate our dataset. 
We invite 12 annotation experts to elaborately annotate each image, and check the annotated images periodically to keep the annotation quality. 
After annotated all images, we conduct a second-round check to remove non-standard ones and select 2,168 usable and well-annotated images from over 4,000 submitted ones.
Fig~\ref{fig::DatasetExamples} shows some examples of SODA.

\begin{table*}[htbp]\scriptsize
\setlength\tabcolsep{2.5pt}
  \centering
\caption{Compare with popular large-scale segmentation datasets.}
    \setlength{\tabcolsep}{0.5mm}
\begin{tabular}{c|c|c|c|c|c|c|cc}
\hline
Name and Reference & Classes & Visible/Depth/Thermal & Resolution  &Samples(training) & Samples(validation) & Samples(test) & Samples(total)\\
\hline 
PASCAL VOC 2012 Segmentation~\cite{everingham2015pascal} & 20 & V & Variable  &1,464 & 1,449 & Private &2,913+\\
PASCAL-Context~\cite{mottaghi2014role} & 540 &  V &  Variable  & 10,103 & N/A & 9,637 &19,740 \\
PASCAL-Part~\cite{chen2014detect} & 20 &  V & Variable &  10,103 & N/A & 9,637 & 19,740 \\
Microsoft COCO~\cite{lin2014microsoft} &80 & V &  Variable &  82,783 & 40,504 & 81,434 & 204,721 \\
Cityscapes(fine)~\cite{cordts2016cityscapes} &30(8)& V &  2048x1024  &2975 & 500 & 1525 & 5,000 \\
CamVid~\cite{brostow2009semantic} &32 & V & 960x720 &701 &N/A &N/A & 701 \\
KITTI-Layout~\cite{alvarez2012road,ros2015unsupervised} &3&V  & Variable &323&N/A & N/A&323 \\
SiftFLow~\cite{liu2009nonparametric} & 33 &V & 256x256 & 2,688 & N/A & N/A & 2,688 \\
NYUDv2~\cite{silberman2012indoor}& 40 & V+D & 480x640 &795 & 654 & N/A & 1449 \\
SUNRGBD~\cite{xiao2013sun3d} & 37 & V+D & Variable & 2,666 & 2,619 & 5,050 & 10,335 \\
RGB-D Object Dataset~\cite{song2015sun} & 51 & V+D & 640x480 & 207,920 & N/A & N/A & 207,903 \\
SODA(+Synthetic Cityscapes fine) &20 & T & 640x480 & 1,168+5,000 & N/A & 1,000 & 7,168 \\
\hline
\end{tabular}%
\label{tab:compare_other_datasets}%
\end{table*}%

For the better understanding our dataset, we analyze the statistics of real part in the SODA dataset. 
There are 20 categories divided into two types: (i) indoor items and (ii) outdoor items. 
\textit{Indoor items} include the common categories such as chair, lamp, monitor, table. 
\textit{Outdoor items} include tree, road, building, sky, car, grass, etc.
%
The characteristics of thermal images are often ambiguity and fuzzy compared to RGB images. 
In total, there are 2,168 images, as shown in Fig~\ref{fig::DatasetStatistics}, including 169 animal images, 392 bicycle images, 1,004 building images, 768 car images, 136 chair images, 52 door images, 153 fence images, 738 grass images, 98 lamp images, 75 monitor images, 1,203 person images, 485 pole images, 122 river images, 1,304 road images, 739 sidewalk images, 830 sky images, 69 table images, 202 trafficCone images, 160 trash can images, as well as 1,519 tree images.

\subsection{Synthetic Subset}
As discussed above, we also construct a set of synthetic thermal images to save manual cost and improve the quality of annotations.

To this end, we use the pix2pixHD~\cite{wang2018high}, a image-to-image translation method, to transform labeled RGB images into thermal images, and the labels of thermal images could directly be transfered from RGB ones.
Among the paired datasets~\cite{Hwang15cvpr,Li16tip,li2018rgb}, the biggest and most relevant is KAIST Multispectral Pedestrian Dataset~\cite{Hwang15cvpr},
which contains a significant amount of aligned images in the RGB and thermal modalities, captured from a moving vehicle in different urban environments and under different lighting conditions. 
We follow the paper~\cite{Hwang15cvpr} and use all the frames of training videos to train the pix2pixHD. 

According the relevance to the thermal images in our real subset, we select the Cityscapes~\cite{cordts2016cityscapes}, an RGB-based cityscapes dataset for semantic urban scene understanding, to generate our synthetic subset.
After acquiring trained model, we translate Cityscapes~\cite{cordts2016cityscapes} to thermal cityscapes, as shown in Fig~\ref{fig::cityscapes}. 
The new generated subset is made up of synthetic thermal images and original semantic label with fine annotations. 
The result shown in~Fig \ref{fig::cityscapes}.

\subsection{Training and Testing Sets}
Considering the scale of our SODA dataset, we divide it into training and testing sets. 
Usually, the proportion of training and testing test sets are about 1:1 in semantic segmentation dataset~\cite{mottaghi2014role,chen2014detect,lin2014microsoft}. 
Therefore, we divide the manual-annotated subset into 1,168 training images and 1,000 testing images. 
We also translate the Cityscapes~\cite{cordts2016cityscapes} dataset into a thermal dataset, and use it to learn parameters of our network for effective training.
Note that the original dataset contains 5,000 images in total, including 2,975 training images, 500 validation images and 1,525 test images.
We take all translated thermal images from this dataset as a part of our training set.
The details of our SODA against other popular datasets are shown in Table~\ref{tab:compare_other_datasets}.

\begin{table*}[htbp]\scriptsize
\setlength\tabcolsep{2.5pt}
  \centering
\caption{Performance comparison in terms of per-clas IoU with eight methods on the SODA test}
    \setlength{\tabcolsep}{0.5mm}
    \begin{tabular}{c|cccccccccccccccccccc|c}
\hline
          & {\bf person} & {\bf building} & {\bf tree} & {\bf road} & {\bf pole} & {\bf grass} & {\bf door} & {\bf table} & {\bf chair} & {\bf car} & {\bf bicycle} & {\bf lamp} & {\bf monitor} & {\bf trafficCone} & {\bf trash can} & {\bf animal} & {\bf fence} & {\bf sky} & {\bf river} & {\bf sidewalk} & {\bf mIoU} \\
\hline
    U-Net\cite{ronneberger2015u} & 31.82 & 37.79 & 22.92 & 60.59 & 3.79 & 21.85 & 0    & 0    & 0   & 26.83 & 5.99  & 4.39  & 26.68 & 5.94  & 0.1   & 0.42  & 21.84    & 34.61 & 25.19 & 16.41 & 19.36 \\
    ERFNet\cite{romera2018erfnet} & 47.31 & 54.86 & 46.37 & 46.25 & 0   & 22.91 & 0     & 0     & 0     & 42.42 & 12.11 & 0     & 28.89 & 0     & 0     & 0       & 0     & 36.25 & 48.17 & 18.35 & 24.69 \\
    FCN-32s\cite{long2015fully} & 27.43 & 38.05 & 47.35 & 56.93 & 0     & 24.66 & 25.51 & 7.38  & 17.49 & 29.32 & 18.97 & 4.67 & 42.33 & 3.91  & 22.61 & 17.07 & 26.53    & 53.05 & 47.14 & 18.2  & 26.43 \\
    FCN-16s\cite{long2015fully} & 40.78 & 45.01 & 48.24 & 63.31 & 18.21 & 26.07 & 0.8   & 0     & 4.49  & 35.55 & 23.44 & 5.93 & 33.89 & 20.33 & 9.03  & 12.35 & 29.68    & 67.49 & 27.13 & 23.05 & 27.24 \\
    DeepLabv2\cite{chen2018deeplab} & 49.81 & 61.32 & 32.24 & 73.36 & 12.27 & 38.67 & 5.99  & 3.86  & 24    & 56.3  & 24.41 & 3.59  & 55.6  & 18.01 & 40.99 & 31.49 & 40.02 &  79.77 & 74.71 & 34.42 & 39.54 \\
    DUC\cite{wang2018understanding}   & \textcolor{blue}{66.14} & \textcolor{blue}{71.75} & \textcolor{green}{76.97} & \textcolor{green}{81.88} & \textcolor{red}{38.4}  & \textcolor{green}{61.54} & 11.31  & 12.24 & 10.55 & 68.55 & \textcolor{blue}{46.34} & \textcolor{blue}{6.43} & 50.51 & \textcolor{blue}{28.48} & 42.99 & \textcolor{blue}{48.9}  & 48.83    & \textcolor{green}{83.74} & 60.47  & \textcolor{green}{49.14} & 49.48 \\
    GCN\cite{peng2017large}   & 64.65 & 68.4  & 68.6  & 76.81 & 21.66 & 48.63 & \textcolor{blue}{16.43} & \textcolor{red}{18.17} & \textcolor{blue}{39.73} & \textcolor{blue}{73.94} & 44.6  & 5.37 & \textcolor{blue}{59.12} & 35.07 & \textcolor{blue}{58.28} & 42.9  & \textcolor{green}{56.31}  & 75.05 & \textcolor{blue}{78.1}  & 46.16 & \textcolor{blue}{51.04} \\
    DeepLabv3\cite{chen2017rethinking} & \textcolor{green}{68.24} & \textcolor{green}{72.49} & \textcolor{blue}{70.97} & \textcolor{blue}{78.78}  & \textcolor{blue}{30.39} & \textcolor{blue}{50.26} &  \textcolor{green}{32.85} & \textcolor{blue}{13.29} & \textcolor{green}{43.22} & \textcolor{red}{76.93} & \textcolor{green}{49.88} & \textcolor{red}{7.73} & \textcolor{green}{61.89} & \textcolor{green}{49.02} & \textcolor{green}{62.38} & \textcolor{green}{72.02}   & \textcolor{blue}{53.20} & \textcolor{blue}{81.55} & \textcolor{green}{78.77} & \textcolor{blue}{48.52} & \textcolor{green}{57.1} \\

\hline
    EC-CNN &\textcolor{red}{71.26} & \textcolor{red}{78.73} & \textcolor{red}{80.81} & \textcolor{red}{84.25} & \textcolor{green}{37.61} & \textcolor{red}{63.25} & \textcolor{red}{43.52} & \textcolor{green}{17.71} & \textcolor{red}{58.72}  & \textcolor{red}{86.04} &\textcolor{red}{58.85}& \textcolor{green}{7.1} &\textcolor{red}{67.05} & \textcolor{red}{58.9} & \textcolor{red}{68.49} & \textcolor{red}{72.08} & \textcolor{red}{58.06} &\textcolor{red}{88.08}    & \textcolor{red}{83.56} & \textcolor{red}{54.13}  & \textcolor{red}{61.9} \\

\hline
    \end{tabular}%

  \label{tab:addlabel}%
\end{table*}%

\section{Performance Evaluation}
Our network is trained on the platform of NVIDIA GeForce GTX TITAN XP GPU with 12GB memory and the implementation is based on the public platform PyTorch.


\subsection{Parameter Setting}
\label{section:experimental settings}
In the generation of thermal images, we train the pix2pixHD~\cite{wang2018high} from scratch, randomly initializing the weights from a Gaussian distribution with the mean of 0 and the standard deviation of 0.02. 
The epochs, batch size and learning rate are set to 20, 1 and 0.0002, respectively. 

For the network branch of semantic segmentation, we use the ``poly'' learning rate policy~\cite{chen2018deeplab} where current learning rate equals to the base one multiplying (1-$\frac{iter} {max\_iter})^{power}$. 
We set the base learning rate to 0.001 and the $power$ to 0.9. 
In addition, we set the iteration number, momentum and weight decay to 100 epochs, 0.9 and 0.0001 respectively. 
The batch size is set to 14 and the size of input images is scaled to 480$\times$480. 
To avoid overfitting in training, we adopt several techniques of data augmentation including mirroring, resizing between 0.5 and 2 and cropping. 
For the branch of the EdgeNet, we first initialize its parameters using the pretrained VGG-16-Net model on the ImageNet dataset, and then fine-tuned it using the BSD500 dataset~\cite{martin2004learning}, where all hyper-parameters are the same with the setting in~\cite{xie2015holistically}.



\begin{table}[]
\caption{Comparison of different initialized model on EC-CNN. The best results are achieved when combing both ImageNet-pretrained and generated TIR model.}
\centering
\scalebox{0.9}{
\begin{tabular}{@{}lccll@{}}
\toprule
Method        & Pixel Acc(\%)& Mean IoU(\%) &  &  \\ \midrule
random initialization    & 63.4        & 24.8          &  &  \\
synthetic thermal model & 72.3         & 36.6          &  &  \\ 
ImageNet model & 83.1         & 60.4          &  &  \\ 
ImageNet model + synthetic thermal model & 83.6        & 61.3          &  &  \\  \midrule
\end{tabular}}
\label{tab:eccn_TIR_ImageNet}
\end{table}

\begin{table}[]
\caption{Result on different stages of EC-CNN. The shadow layers have better result.}
\begin{tabular}{@{}clllcc@{}}
\toprule
model & conv2\_x  & conv3\_x & conv4\_x &  Pixel Acc(\%) & Mean IoU(\%) \\ \midrule
\multicolumn{6}{l}{Baseline:DeepLabv3}                       \\ \midrule
50 & & &  \multicolumn{1}{l|}{}       &\textcolor{blue}{79.9}         & \textcolor{blue}{57.1}        \\
50    & \checkmark     &       & \multicolumn{1}{l|}{}       & \textcolor{red}{83.6}         & \textcolor{red}{61.3}         \\
50    &       & \checkmark     & \multicolumn{1}{l|}{}       &\textcolor{green}{82.2}         & \textcolor{green}{58.8}          \\
50    &       &       &  \multicolumn{1}{l|}{\checkmark}     & 78.1        & 49.2          \\
50    & \checkmark     & \checkmark     & \multicolumn{1}{l|}{}      & 82.5        & 58.0          \\
50    &       & \checkmark     &\multicolumn{1}{l|}{\checkmark}   & 77.6         & 45.9          \\
50    & \checkmark     &       & \multicolumn{1}{l|}{\checkmark}     &79.1        & 49.3          \\
50    & \checkmark    & \checkmark     & \multicolumn{1}{l|}{\checkmark}    & 77.9         & 46.7          \\ \midrule
101    &      &       &\multicolumn{1}{l|}{}       &81.0        &59.5         \\ 
101    & \checkmark     &       &\multicolumn{1}{l|}{}       &\textcolor{green}{83.8}        &\textcolor{green}{61.7}         \\
101    &       & \checkmark     &  \multicolumn{1}{l|}{}     &\textcolor{red}{83.9}        &\textcolor{red}{61.9}         \\
101   &       &       & \multicolumn{1}{l|}{\checkmark}    &76.9         &45.8          \\
101    & \checkmark     & \checkmark     &  \multicolumn{1}{l|}{}     &\textcolor{blue}{83.5}        & \textcolor{blue}{61.0}          \\
101    &       & \checkmark     & \multicolumn{1}{l|}{\checkmark}     &68.9         &35.5          \\
101    & \checkmark     &       &\multicolumn{1}{l|}{\checkmark}   &77.8         &49.0          \\
101    & \checkmark    & \checkmark     & \multicolumn{1}{l|}{\checkmark}    &70.7         &37.3 \\ \bottomrule
\end{tabular}
\label{tab:eccn_result}
\end{table}

\begin{table}[]
\caption{Result of Baseline, SFT and our GFT respectively..}
\centering
\begin{tabular}{@{}lccll@{}}
\toprule
Method        & Pixel Acc(\%) & Mean IoU(\%) &  &  \\ \midrule
Baseline(50)    & 79.9        &57.1        &  &  \\
SFT(conv2\_x+50)  & 82.3         & 60.8          &  &  \\
GFT(conv2\_x+50) & 83.6         & 61.3         &  &  \\ \bottomrule
\end{tabular}
\label{tab:v3_sft_gft}
\end{table}

\subsection{Evaluation Metric}
\label{section:eval}
We use two standard evaluation metrics including the pixel accuracy (pixAcc) and the mean Intersection of Union (mIoU). 
Note that we use the VOC-like evaluation server that calculates mIoU considering the background as one of the categories.

\subsection{Overall Performance}
\label{section:performance evaluation}
We first compare the proposed approach with the state-of-the-art methods on our SODA dataset.

We report the results with eight state-of-the-art methods including U-Net~\cite{ronneberger2015u}, ERFNet~\cite{romera2018erfnet}, FCN-16s~\cite{long2015fully}, FCN-32s~\cite{long2015fully}, Deeplabv2~\cite{chen2018deeplab}, DUC~\cite{wang2018understanding}, GCN~\cite{peng2017large}, DeepLabv3~\cite{chen2017rethinking}, on SODA in Table~\ref{tab:addlabel}. 
Herein, we utilize the ResNet-101 network as the backbone, and apply GFT layers in conv3\_x. 
For ERFNet and GCN, we re-implemented their methods and fine-tuned their models with the same setting as ours. 
For other methods, we directly use their released codes.

One can see that, our method achieves 61.9\% in mean IoU, better than DeepLabv3~\cite{chen2017rethinking} with 4.8\% gain and GCN~\cite{peng2017large} with 10.86 gain, which are the second and third best methods. 
This superior performance achieved by our method demonstrates the effectiveness of the proposed EC-CNN.
We further report per-class IoU on SODA to verify the class-based performance of EC-CNN, presented in Table~\ref{tab:addlabel}, and we achieve the best performance on a majority of classes. 

\begin{figure*}[t]
  \centering
  \includegraphics[width=\textwidth]{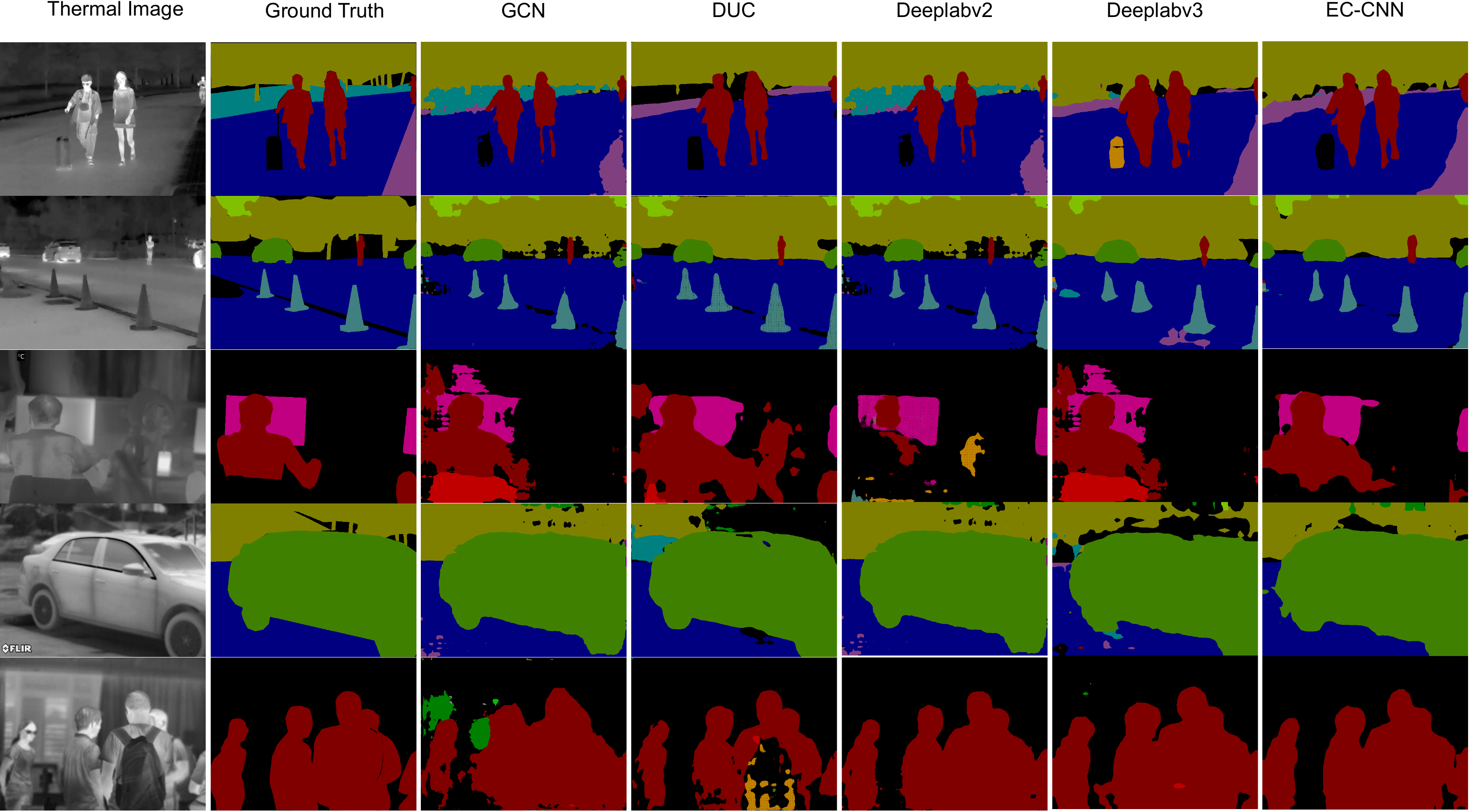} \\
  \caption{Examples of thermal semantic segmentation results on SODA.  }\label{fig::compare_result}
\end{figure*}

The qualitative comparison is visualized in Fig.~\ref{fig::compare_result}. 
As can be observed from these visual results, our method outputs more precise predictions than other four methods despite of the existence of low resolution and contrast, ambiguous object boundaries caused by thermal crossover, and imaging noised introduced by thermal sensors. 
Taking the third and fifth row for examples, our approaches can also successfully handle the ambiguous object boundaries caused by thermal crossover, such as the left arm and TV monitor. 
These regions with similar textures can be recognized and separated by the guidance from the introduced edge information. 
In general, through effectively exploiting edge priors, our approach produces more reasonable results of thermal image semantic segmentation in spite of the presences of confusing labels.

\subsection{Ablation Study}
\label{section:ablation studies}

\begin{table}[]
\caption{Speed grows with deeper networks. Number in the brackets refers to the depth of ResNet and `conv2\_x' denotes add edge information in the conv2\_x of EC-CNN.}
\centering
\begin{tabular}{@{}lccll@{}}
\toprule
Method        & time &  &  \\ \midrule
Deeplabv3(50)   &  0.11    &  &  \\
EC-CNN(conv2\_x+50)     & 0.13      &  &  \\
EC-CNN(conv3\_x+50)    &  0.14    &  &  \\
EC-CNN(conv4\_x+50)  & 0.17        &  &  \\ \midrule
Deeplabv3(101)     &0.18       &  &  \\
EC-CNN(conv2\_x+101)     & 0.20       &  &  \\
EC-CNN(conv3\_x+101)    & 0.21       &  &  \\
EC-CNN(conv4\_x+101)  &  0.45    &  &  \\ \bottomrule
\end{tabular}
\label{tab:inf_speed}
\end{table}

{\flushleft \bf Quality of thermal image generation.}
To evaluate the visual quality of thermal data in synthetic subset, we evaluate the visual satisfaction of the translated thermal images.
Some examples are shown in Fig.~\ref{fig::pix2pixHD_pix2pix_demo}, where the results of pix2pix~\cite{isola2017image} and the corresponding ground-truths are presented.
We can see that the pix2pixHD technique~\cite{wang2018high} adopted in our work can generate better thermal images from RGB ones.
It validates the feasibility of the integration of the generated thermal images in our training set.

{\flushleft \bf Impact of synthetic thermal data.}
To further justify the effectiveness of our synthetic thermal data, we design four different configurations of our EC-CNN. 
They are: 1) random initialization, that randomly initializes network parameters in EC-CNN; 
2) synthetic thermal model, that only uses the synthetic thermal subset to pre-train EC-CNN;
3) ImageNet model, that only uses the ImageNet dataset to pre-train EC-CNN;
and 4) ImageNet model + synthetic thermal model, that first employs the ImageNet dataset to pre-train EC-CNN and then synthetic thermal subset.

The evaluation results are reported in the Table~\ref{tab:eccn_TIR_ImageNet}.
Comparing the results of ``synthetic thermal model'' with ``random initialization'', Pixel Acc rises from 63.4\% to 72.3\% and mIoU rises from 24.8\% to 36.6\%.
Moreover, the results of ``ImageNet model + synthetic thermal model'' also outperforms ``ImageNet model'' on both .
These observations suggest that the incorporation of synthetic thermal data is beneficial to improving the performance of thermal image semantic segmentation. 
In addition, the results of ``ImageNet model + synthetic thermal model'' are best, which justifies the effectiveness of our strategy for network training.

{\flushleft \bf Impact of GFT layer.} 
We apply the GFT layer in different convolutional blocks including conv2\_x, conv3\_x, conv4\_x and combination of them, to find a best scheme.
Table~\ref{tab:eccn_result} shows the details of configurations and the evaluation results on both ResNet-50 and ResNet-101 backbones.
We can see that the GFT layers are inserted in shadower blocks, the better results are achieved.
For example, the result of 50 with conv2\_x, which means inserting GFT layers into conv2\_x block with ResNet-50 as the backbone, is best and clearly outperform 50 with conv3\_x or conv4\_x.
Furthermore, the performance of 50 with conv2\_x and conv3\_x is better than 50 with conv2\_x and conv4\_x, and both of them are better than 50 with conv3\_x and conv4\_x. 
The similar observations are obtained on the ResNet-101 backbone. 
It can be explained by the fact that the edge prior information is low-level features and incorporating them into shadow blocks would lead to a better embedding in thermal image semantic segmentation.

{\flushleft \bf Impact of gate mechanism.}
To validate the effectiveness of the gate mechanism in the proposed GFT layer, we remove it and denote the layer as SFT layer, and report the evaluation results in Table~\ref{tab:v3_sft_gft}.
Comparing with SFT, we obtain an improvement of 1.3\% (from 82.3\% to 83.6\%) in Pixel Acc and 0.5\% (from 60.8\% $\to$ 61.3\%) in mIoU. 
This indicates through adding the gate mechanism in the GFT layer, we can alleviate noise effects introduced by EdgeNet and prefer to use good edge information to guide semantic segmentation.

{\flushleft \bf Comparison of inference speed.}
We finally evaluate the inference speed of the proposed framework. As in Table~\ref{tab:inf_speed}, the EC-CNN (ResNet50) takes about 0.13-0.17s/image in average on GeForce GTX TITAN XP GPU, which are near to the speed of baseline.
The results also have similar time consuming observations on ResNet-101. 
It suggests that our framework can achieve clear improvements in both pixAcc and mIoU over the baseline DeepLabv3 with a modest impact on the inference speed.

\section{Concluding Remark}
In this work, we presented a newly dataset SODA for a new vision task of thermal image semantic segmentation. 
SODA contains 2,168 images, which are richly annotated with 20 semantic labels.
Besides, we also provide Cityscapse-based synthetic thermal images to increase the scale of SODA. 
With the SODA dataset, we performed extensive experimental analysis to identify successes and limitations of some leading approaches on the task of thermal image semantic segmentation, and proposed a more suitable method for this task.
In a particular, we design a novel EC-CNN architecture to explicitly incorporate edge priors.
The experimental results demonstrate the effectiveness of the proposed EC-CNN.

From the evaluation results, utilizing edge prior information in the task of thermal image semantic segmentation has many advantages especially when semantic regions are low resolution and contrast, boundary-ambiguous caused by thermal crossover, and imaging noises introduced by thermal sensors.

In addition, based on evaluation results and observations, we highlight some components which are essential for thermal image semantic segmentation. 
The first one is the dataset. 
At present, there is no large-scale annotated dataset for thermal image semantic segmentation.
Although our SODA provide a first dataset for this new task, the well-annotated data are not enough, especially for the development of deep learning techniques in the research direction.
Therefore, a large-scale well-annotated thermal image dataset is essential for boosting development of thermal image semantic segmentation.
Meanwhile, we should also study transfer learning techniques to transform massive knowledge from RGB data into thermal domain which alleviates the dependence on large-scale data. 

The second one is the method.
As discussed above, thermal information has special properties comparing with visible spectrum data (i.e., RGB data), such as insensitivity to lighting conditions and thermal crossover.
How to design a suitable and powerful network for thermal data in extremely important. 
We provide one solution by incorporating edge priors in existing RGB semantic segmentation network, and other priors can also be investigated, such as semantic relations and context knowledge. 
In addition, we also use the advantages of RGB data to mitigate the disadvantages of thermal data like thermal crossover using multi-modal learning techniques. 
On one side, the thermal sensors are more effective in capturing objects than visible spectrum cameras under poor lighting conditions and bad weathers. 
On the other side, visible spectrum cameras are more discernible in separating two different subjects under thermal crossover. 
Therefore, RGB and thermal information complement each other and contribute to semantic segmentation in different aspects together.

\bibliographystyle{IEEEtran}
\bibliography{IEEEref}




%




\end{document}